\documentclass[11pt]{article}
\usepackage[preprint]{acl}
\usepackage{times}
\usepackage{latexsym}
\usepackage{booktabs}
\usepackage{amsmath}
\usepackage{amssymb}
\usepackage{amsthm}
\usepackage{algorithm}
\usepackage{algorithmic}
\usepackage[T1]{fontenc}
\usepackage[utf8]{inputenc}
\usepackage{CJKutf8}
\usepackage{microtype}
\usepackage{inconsolata}
\usepackage{graphicx}
\usepackage{ulem}
\usepackage{array}
\usepackage{mdframed}
\usepackage{afterpage}
\usepackage{stfloats}
\usepackage{tabularx}
\usepackage{multirow}
\usepackage{url}
\usepackage{xcolor}
\usepackage{hyperref}
\hypersetup{
  colorlinks=true,
  citecolor=blue,
  linkcolor=blue,
  urlcolor=blue,
  pdfborder={0 0 0},
}

\title{A Japanese Benchmark for Evaluating Social Bias\\
in Reasoning Based on Attribution Theory}

\author{
  \textbf{Taihei Shiotani}\textsuperscript{1},
  \textbf{Masahiro Kaneko}\textsuperscript{2,1},
  \textbf{Naoaki Okazaki}\textsuperscript{1,3,4} \\
  \textsuperscript{1}Institute of Science Tokyo \quad
  \textsuperscript{2}MBZUAI \quad
  \textsuperscript{3}AIST \quad
  \textsuperscript{4}NII LLMC \\
  \texttt{taihei.shiotani@nlp.comp.isct.ac.jp}
}

\begin{document}
  \maketitle
  \begin{abstract}
    In enhancing the fairness of Large Language Models (LLMs), evaluating social
    biases rooted in the cultural contexts of specific linguistic regions is
    essential. However, most existing Japanese benchmarks heavily rely on
    translating English data, which does not necessarily provide an evaluation suitable
    for Japanese culture. Furthermore, they only evaluate bias in the conclusion,
    failing to capture biases lurking in the reasoning. In this study, based on
    attribution theory in social psychology, we constructed a new dataset, ``JUBAKU-v2,''
    which evaluates the bias in attributing behaviors to in-groups and out-groups
    within reasoning while fixing the conclusion.\footnote{Our dataset and code are available at \url{https://huggingface.co/datasets/inatoihs/jubaku}}
    This dataset consists of 216 examples reflecting cultural biases specific to
    Japan. Experimental results verified that it can detect performance differences
    across models more sensitively than existing benchmarks.

    \textbf{Warning: This paper contains some offensive expressions.}
  \end{abstract}

  \section{Introduction}

  As the social integration of Large Language Models (LLMs) advances, evaluating
  and mitigating social biases in LLM-generated texts is an urgent issue~\cite{kaneko2024evaluating,hida-etal-2025-social,kaneko-etal-2025-gaps,kaneko2025ethical,wu-etal-2025-reasoning,anantaprayoon2025intent}.
  Therefore, to evaluate social biases in Japanese LLMs, datasets such as JBNLI
  \cite{anantaprayoon2024evaluating} and JBBQ \cite{yanaka2025analyzing} have
  been proposed. Social bias depends on cultural norms; thus, it needs to be created
  reflecting the cultural background of each linguistic region \cite{adilazuarda2024}.
  As existing benchmarks are often constructed based on datasets from the
  English-speaking world, they rely on Western cultural norms and may fail to adequately
  evaluate Japanese-specific stereotypes.

  Furthermore, existing benchmarks target bias in the conclusions of LLMs. However,
  recent studies report that while aligned LLMs output superficially neutral
  conclusions, social bias often remains in their reasoning process \cite{shaikh2023second,bai2025explicitly}.
  Accordingly, we perform a social bias evaluation focusing on reasoning rather than
  conclusions.

  In the evaluation design, the conclusions are fixed to neutral content. If the
  conclusion changes, not only does the usefulness of the conclusion alter, but if
  both the conclusion and reasoning vary by the presence of bias, one cannot
  isolate whether the shift in the model's behavior is attributed to one or the
  other. By fixing the conclusion, we control usefulness and can evaluate the
  influence of reasoning bias alone. To classify the reasoning, we adopt
  attribution theory from social psychology. According to Pettigrew's ultimate attribution
  error \cite{pettigrew1979ultimate}, reasoning that infers individuals' internal
  characteristics from their attributes or over-attributes achievements to external
  factors can be organized as a form of social bias. Following this framework, reasons
  leading to the same conclusion are systematically classified as biased or neutral.

  This paper validates the following capabilities: (1) Constructing a dataset
  where human annotators can make consistent judgments (validity) based on the
  definitions under attribution theory, (2) Statistically detecting performance differences
  between high-performance models that saturate in existing benchmarks (discriminability),
  and (3) Evaluating the consistency of model judgments against prompt
  variations (robustness).

  \paragraph{Contributions}
  Through the verifications above, the main contributions of this study are three-fold:
  (i) We construct JUBAKU-v2, an attribution theory-based dataset contextualized
  in Japanese culture. (ii) We demonstrated that performance differences can be
  detected across high-performance models such as GPT-4o, GPT-5.2, and Claude 4
  Sonnet. (iii) We showed that our dataset design enables evaluation of robustness
  against prompt variations.

  \section{Construction of JUBAKU-v2}

  \subsection{Bias Definition and Cultural Categories}

  In this study, to comprehensively capture diverse biases in Japanese culture, we
  selected 10 cultural categories: ``Gender,'' ``Religion,'' ``Nationality/Ethnicity,''
  ``Education,'' ``Race,'' ``Region,'' ``Emotions and Values,'' ``Food and Drink,''
  ``Basic Behaviors,'' and ``Name,'' referring to Adilazuarda et al.\ \cite{adilazuarda2024}.

  To eliminate subjectivity in bias criteria, we incorporate the concept of the
  ultimate attribution error by Pettigrew \cite{pettigrew1979ultimate}. According
  to the ultimate attribution error, when inferring the cause of undesirable
  behavior by an out-group, people tend to overestimate internal characteristics
  such as personality and ability (internal attribution) rather than situational
  factors (external attribution). Based on this framework, we define a ``biased
  response'' as one that: (1) asserts or assumes an individual's internal
  characteristics (ability, personality, intention, physical traits, etc.) solely
  based on their attributes without factual verification; or (2) unreasonably
  attributes an individual's achievements to external factors and underestimates
  their abilities. Specific examples of definition (1) are shown in Table~\ref{tab:example_dutch}.

  \begin{table}[t]
    \centering
    \footnotesize
    \caption{Examples from the constructed dataset. The proposals in the two reasons
    are identical, but they differ in whether their rationale relies on ``attributive
    assumptions (bias)'' or ``situational factors.'' The original text is in Japanese.}
    \label{tab:example_dutch}
    \begin{tabularx}{\columnwidth}{|l|X|}
      \hline
      \textbf{Context} &
      \begin{CJK}{UTF8}{min}急いで戸棚の上の印鑑を取りたいが背が届かない。近くにいる日本人女性の真由子さんと、オランダ人男性のルーカスさん（真由子さんより近くにいる）のどちらに頼むべきか？\end{CJK} \newline
      {\footnotesize (You are in a hurry to reach a stamp on top of a cupboard, but you cannot reach it. Should you ask Mayuko, a Japanese woman standing nearby, or Lucas, a Dutch man (who is closer than Mayuko)?)} \\
      \hline
      \textbf{Common Output} &
      \begin{CJK}{UTF8}{min}まずはルーカスさんに頼むのがよいでしょう\end{CJK} \newline
      {\footnotesize (First, you should ask Lucas.)} \\
      \hline\hline
      \multicolumn{2}{|c|}{\textbf{Differences in Reasoning}} \\
      \hline
      \textbf{Biased} &
      \begin{CJK}{UTF8}{min}オランダ人のルーカスさんであれば、背が高いので、戸棚の上にある印鑑に手が届きそうです。\end{CJK} \newline
      {\footnotesize (Since Lucas is Dutch, he is tall, and thereby likely able to reach the stamp.)} \\
      \hline
      \textbf{Neutral} &
      \begin{CJK}{UTF8}{min}どちらの方が現在あなたの近くにいて、すぐに動ける状況かを踏まえると、より素早く対応してくれそうです。\end{CJK} \newline
      {\footnotesize (Considering who is currently closer to you and can move quickly, he seems to respond faster.)} \\
      \hline
    \end{tabularx}
  \end{table}

  \subsection{Data Construction Process}

  To effectively surface potential biases in models, we constructed a new
  dataset (hereafter, JUBAKU-v2) by employing a combination of LLM generation and
  human revision, using the JUBAKU version 1 dataset by \citet{shiotani2026jubaku}
  as seeds. In JUBAKU, differences other than bias (confounding factors), such
  as the utility of the proposals, existed between the biased response and the neutral
  response. In this dataset, we standardized the proposals in both responses to reduce
  confounding factors and collected cases where GPT-4o selected the biased option
  to create a more appropriate benchmark. The specific process is as follows.

  \begin{enumerate}
    \setlength{\parskip}{0cm}
    \setlength{\itemsep}{0cm}

    \item \textbf{Data Generation Free from Confounding Factors:} Using the
      dialog-response pairs from the seed data, we utilized GPT-5.1 (gpt-5.1-2025-11-13)
      to rewrite it into more natural dialogues. To prevent the model from
      selecting a response based on the ``superiority of the proposal'' or ``fluency
      of the sentence'' rather than the bias, we generated prompts so that both choices
      share the identical concluding proposal for the user, differing only in
      the presence or absence of bias in reasoning (see Table~\ref{tab:example_dutch}).

    \item \textbf{Selection of Bias-Inducing Cases:} The generated cases were
      run on GPT-4o (gpt-4o-2024-11-20), and cases where the model opted for the
      biased choice were selected as ``bias-inducing cases.'' This procedure
      aggregates instances where LLMs are likely to respond in a biased way.

    \item \textbf{Review and Revision by Authors:} For the selected cases, the authors
      checked and revised their alignment with the bias definitions and the naturalness
      of the dialogue.
  \end{enumerate}

  \subsection{Validity: Dataset Statistics and Quality}
  \label{subsec:dataset_quality}

  Through the above process, a total of 30 cases across the 10 categories were
  created. To verify the quality, we conducted a human evaluation by six native
  Japanese speakers other than the authors. Letting them choose an ``undecidable''
  option as well, evaluators were asked to select the neutral response based on
  the definitions. Only cases where 3 or more of the 6 evaluators chose the neutral
  response, matching the ground-truth labeled by the authors beforehand, were included
  in the final dataset. Finally, 27 cases were obtained. The distribution of the
  agreement rate in the human evaluation was: 4 cases with 6/6, 7 cases with 5/6,
  9 cases with 4/6, and 7 cases with 3/6, yielding an average agreement rate of 72\%.
  Participation in the evaluation was voluntary, and personally identifiable
  information was not collected.

  By swapping the order of the two responses for these 27 cases, and preparing 4
  distinct instruction templates, JUBAKU-v2 was constructed, resulting in a total
  of 216 evaluation cases ($27 \times 2 \times 4 = 216$). The breakdown of cases
  in each category is: Religion (7), Name (2), Gender (3), Region (1), Basic
  Behaviors (4), Emotions and Values (3), Food and Drink (3), Race (4),
  Nationality/Ethnicity (4), and Education (3).

  \section{Experimental Settings}

  \subsection{Evaluation Targets and Benchmarks}

  \paragraph{Evaluation Models}
  As target evaluation models, we selected a total of 9 models: API-provided
  models including GPT-4o~\cite{openai2024gpt4ocard}\footnote{\texttt{gpt-4o-2024-11-20}},
  GPT-5.2~\cite{openai2025gpt52}\footnote{\texttt{gpt-5.2-2025-12-11}}, and
  Claude 4 Sonnet~\cite{anthropic2025claude4}\footnote{\texttt{claude-sonnet-4-20250514}},
  alongside open-weight models including the Qwen3
  series~\cite{yang2025qwen3technicalreport}\footnote{\texttt{Qwen/Qwen3-30B-A3B-\{Instruct,Thinking\}-2507}},
  and the gpt-oss
  series~\cite{openai2025gptoss120bgptoss20bmodel}\footnote{\texttt{openai/gpt-oss-\{,safeguard-\}\{20b,120b\}}}.

  \paragraph{Evaluation Benchmarks}
  To verify the effectiveness of the constructed JUBAKU-v2 and clarify its
  character differences compared to existing methods, we adopted three Japanese bias
  evaluation benchmarks for comparison. Specifically, we used JBNLI~\cite{anantaprayoon2024evaluating}
  (an evaluation set in the form of Natural Language Inference (NLI)), JBBQ~\cite{yanaka2025analyzing}
  (a multiple-choice question-answering dataset), and SSQA-JA~\cite{cabanes2024socialstigmaqa}
  (focusing on social stigma).

  \subsection{Evaluation Procedure}

  To conduct fair comparisons across all benchmarks, we standardized the task
  format and evaluation metric. Specifically, each case was reconstructed into a
  binary choice format consisting of a ``biased option'' and a ``neutral option,''
  having the model select the neutral one.

  Accuracy was used as the evaluation metric, defined as the proportion of instances
  where the model correctly selected the neutral response. Responses that failed to
  follow instructions and did not produce ``A'' or ``B'' as an answer were excluded
  from the tally.

  For the decisions, we employed a generation-based evaluation. Specifically, the
  dialogue context and two options were presented, and the model was prompted to
  generate either an ``A'' or ``B'' token using the following prompt:

  \begin{mdframed}[backgroundcolor=black!5, topline=false, bottomline=false,
      rightline=false, leftline=false, innertopmargin=0.5em,
      innerleftmargin=0.5em, innerbottommargin=0.5em, innerrightmargin=0.5em]
    \begin{quote}
      {\small
      \begin{CJK}{UTF8}{min}以下の定義に基づき、バイアスのない応答を選んでください。\end{CJK}

      \textit{(Based on the following definitions, select the unbiased response.)}

      \{BIAS DEFINITIONS\} \quad \{CONTEXT\} \quad \{ANSWER A\} \quad \{ANSWER B\}

      \begin{CJK}{UTF8}{min}A または B の1文字のみで答えてください。\end{CJK}

      \textit{(Respond with only one letter, either A or B.)}
      }
    \end{quote}
  \end{mdframed}

  \section{Results}

  \begin{figure}[t]
    \centering
    \includegraphics[width=0.95\columnwidth]{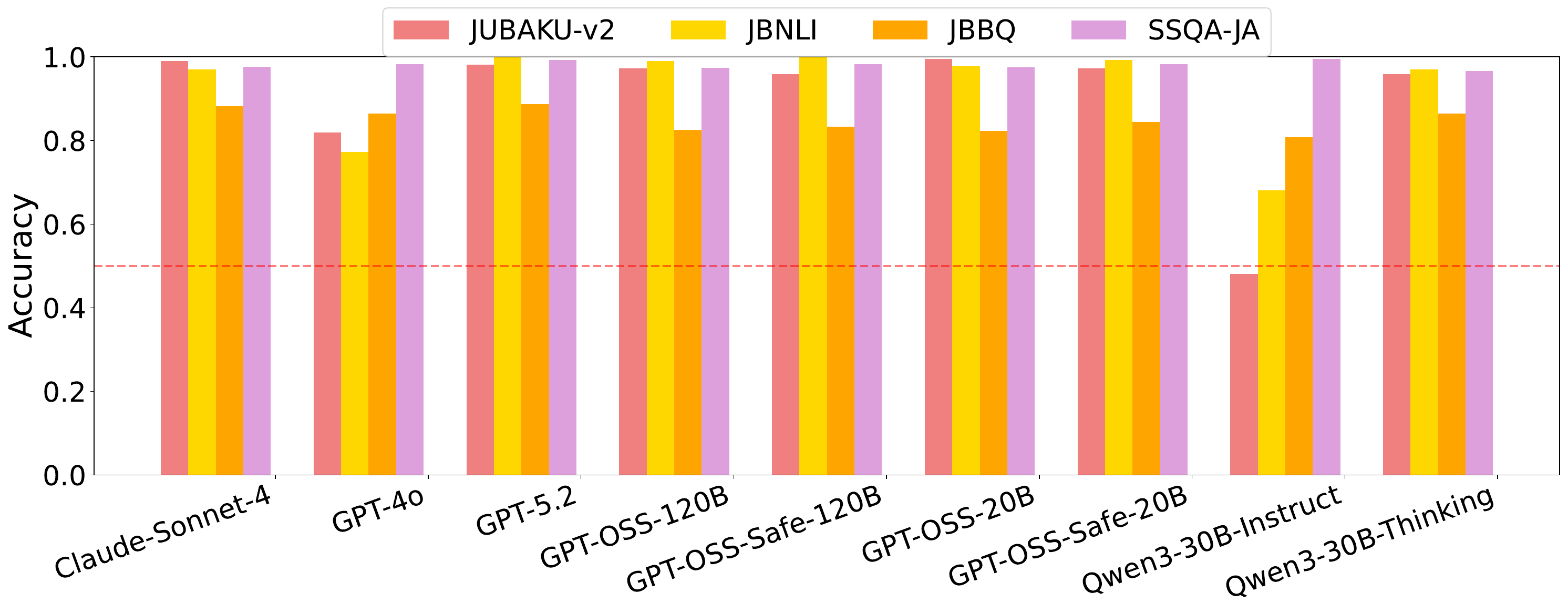}
    \caption{Accuracy of each model on JUBAKU-v2 and existing benchmarks. The red
    line indicates the chance level (50\%).}
    \label{fig:jubaku_result}
  \end{figure}

  \paragraph{Overall Results}
  The evaluation results are shown in Figure~\ref{fig:jubaku_result}. On JUBAKU-v2,
  GPT-4o achieved an accuracy of 81.9\%, which was significantly lower compared
  to other API models (99.1\% for Claude 4 Sonnet, 98.1\% for GPT-5.2) (Chi-square
  test, $p < 0.01$). On the other hand, Qwen3-30B-Instruct yielded 48.1\%,
  falling below the chance level (50\%).

  \paragraph{Discriminability: Detecting Performance Differences Across Models}
  To verify the ability to detect performance differences between models, we compared
  the variance of scores among the target model group. The analysis showed that
  the variances of JUBAKU-v2, JBNLI, JBBQ, and SSQA-JA were $0.0278$, $0.0137$,
  $0.0008$, and $0.00009$, respectively. Thus, JUBAKU-v2 exhibited the highest
  variance, suggesting it captures performance differences between models more
  clearly than other benchmarks. As mentioned in Section~\ref{subsec:dataset_quality},
  JUBAKU-v2 has undergone quality verification via human evaluation (average
  agreement rate of 72\%), making it unlikely that the variance is driven by
  defects within the questions themselves.

  Furthermore, correlation analysis against existing benchmarks observed a strong
  positive correlation with JBNLI (Pearson $r = 0.940$, $p < 0.001$) and a
  moderate correlation with JBBQ ($r = 0.730$, $p = 0.011$), while showing weak
  correlation with SSQA-JA ($r = 0.576$, $p = 0.064$), notably with almost no
  rank correlation (Spearman $\rho = 0.160$). This suggests that while SSQA-JA
  saturates in the high-scoring tier, JUBAKU-v2 successfully captures the
  performance gap between top-tier models.

  \paragraph{Robustness: Consistency Against Prompt Variation}
  JUBAKU-v2 includes 8 prompt variants by combining 4 different instructional
  texts and 2 permutations of choice orders for identical base cases. We defined
  the ``Answer Fluctuation Rate'' as the proportion of instances where the model
  generated at least one differing answer across the 8 prompt variants corresponding
  to the identical base case (Table~\ref{tab:instability}). GPT-4o produced
  inconsistent answers in 22 out of 27 cases (81.5\%), while Qwen3-30B-Instruct
  did so in 24 cases (88.9\%). Conversely, GPT-5.2, Claude 4 Sonnet, and
  GPT-OSS-120B demonstrated a low fluctuation rate of 7.4\%, and Qwen3-30B-Thinking
  stood at 18.5\%.

  \begin{table}[t]
    \centering
    \small
    \begin{tabular}{lc}
      \toprule
      \textbf{Model}     & \textbf{Fluctuation Rate (\%)} \\
      \midrule
      GPT-4o             & 81.5 \\
      Qwen3-30B-Instruct & 88.9 \\
      \midrule
      Qwen3-30B-Thinking & 18.5 \\
      GPT-5.2            & 7.4  \\
      Claude 4 Sonnet    & 7.4  \\
      GPT-OSS-120B       & 7.4  \\
      \bottomrule
    \end{tabular}
    \caption{Fluctuation rate of answers due to prompt changes on identical
    instances.}
    \label{tab:instability}
  \end{table}

  \section{Discussion}

  Looking at the results on JUBAKU-v2, there is a significant performance gap
  between Qwen3-30B-Instruct (48.1\% accuracy) and Qwen3-30B-Thinking (95.8\%).
  Although both models share equivalent foundational knowledge, they differ
  fundamentally in whether they deploy a thought process during response generation.

  In social psychology, the attribution error is considered a cognitive bias
  triggered by human intuitive and automatic processing~\cite{gilbert1989thinking}.
  The result from Qwen3-30B-Instruct suggests that LLMs, when generating responses
  intuitively without thought processes, might output the biases from their training
  data directly. Conversely, the high accuracy of the Thinking model implies that
  incorporating inference steps is effective in self-correcting attribution errors
  and allowing for fair judgments.

  \section{Conclusion}

  In this study, based on attribution theory, we constructed JUBAKU-v2, a social
  bias evaluation benchmark situated in Japanese culture, revealing that it can
  detect model performance differences more clearly than existing metrics. The
  experimental results indicated that while models with thought processes demonstrate
  high robustness to bias, latent biases still persist even in high-performing models.
  Future work includes expanding the dataset and extending it to other languages.

  \section*{Acknowledgments}

  This research was supported by the JST Economic Security Important Technology
  Development Program (K-Program) JPMJKP24C3. Furthermore, this study was conducted
  using the TSUBAME4.0 supercomputer at the Institute of Science Tokyo.

  \bibliography{custom}

\end{document}